\documentclass{article}
\usepackage{spconf,amsmath,graphicx}
\usepackage{booktabs}

\title{Query-Utterance Attention with Joint modeling for Query-Focused Meeting Summarization}
\name{Xingxian Liu, Bin Duan, Bo Xiao, \thanks{Yajing Xu is the corresponding author.}Yajing Xu \thanks{This work was supported by MoE-CMCC "Artifical Intelligence" Project No. MCM20190701 and the National Natural Science Foundation
of China (NSFC No.62076031).}}
\address{Beijing University of Posts and Telecommunications, Beijing, China\\}

\begin{document}
%
\maketitle
\begin{abstract}
Query-focused meeting summarization (QFMS) aims to generate summaries from meeting transcripts in response to a given query.
Previous works typically concatenate the query with meeting transcripts and implicitly model the query relevance only at the token level with attention mechanism.
However, due to the dilution of key query-relevant information caused by long meeting transcripts, the original transformer-based model is insufficient to highlight the key parts related to the query. 
In this paper, we propose a query-aware framework with joint modeling token and utterance based on Query-Utterance Attention.
It calculates the utterance-level relevance to the query with a dense retrieval module.
Then both token-level query relevance and utterance-level query relevance are combined and incorporated into the generation process with attention mechanism explicitly.
We show that the query relevance of different granularities contributes to generating a summary more related to the query.
Experimental results on the QMSum dataset show that the proposed model achieves new state-of-the-art performance.
\end{abstract}
\section{Introduction}

Meeting summarization aims to summarize the key parts of a meeting transcript into a concise textual passage to improve the efficiency of access to information.
People are usually more interested in certain parts than the whole meeting.
Query-focused meeting summarization (QFMS) adopts a more interactive approach that accepts user's query and generates a summary in response to the query \cite{litvak-vanetik-2017-query, baumel2018query}.
Therefore, QFMS can satisfy user's need to focus on specific topics and aspects of the meeting.

Early attempts \cite{laskar-etal-2020-wsl, xu-lapata-2020-coarse, xu-lapata-2021-generating} in the query-focused summarization (QFS) focused on alleviating the data deficit problem.
Early work \cite{baumel2018query} in multi-document summarization models sentence-query score with token-level overlap and introduces the score in the RNN-based model.
Recently, introduction of the high-quality dataset QMSUM \cite{zhong-etal-2021-qmsum} has significantly advanced the research of QFMS.

\begin{figure}[t]
	\centering
    \includegraphics[scale=0.60]{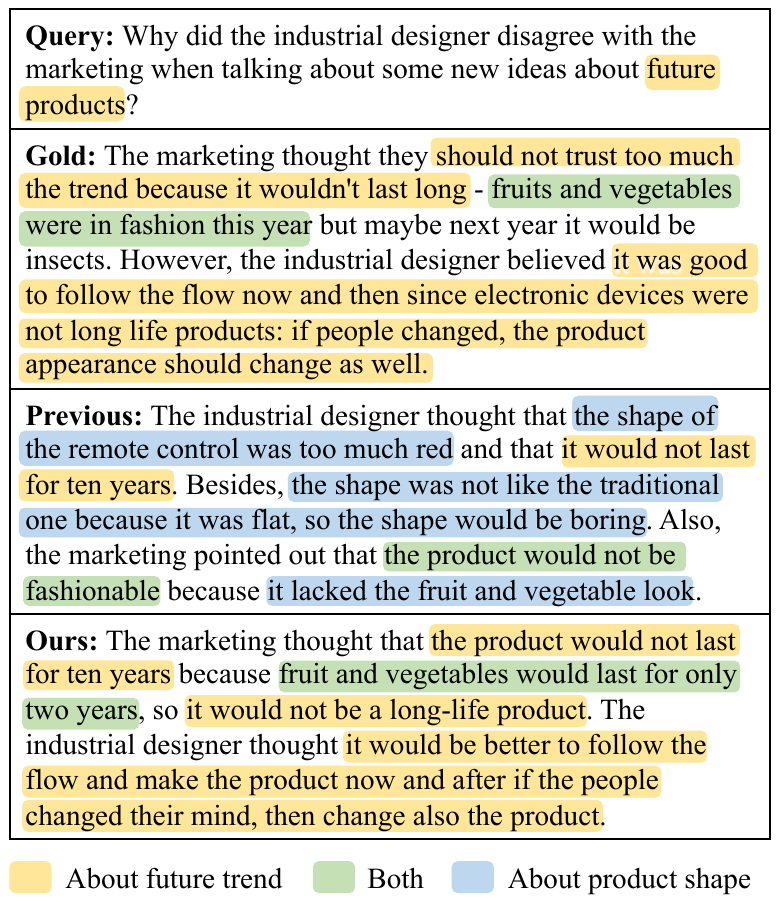}%
    \vspace{-0.15cm}
	\caption{
	Different topics are marked with different colors. 
	Without explicit modeling Query-Utterance relevance, the previous model generated summaries that deviated from the query's topic. 
	In contrast, by explicitly modeling Query-Utterance relevance, our model can better capture the topics of the query.
	}
	\label{fig1}
	\vspace{-0.4cm}
\end{figure}

\begin{figure*}[t]
	\centering
    \includegraphics[scale=0.80]{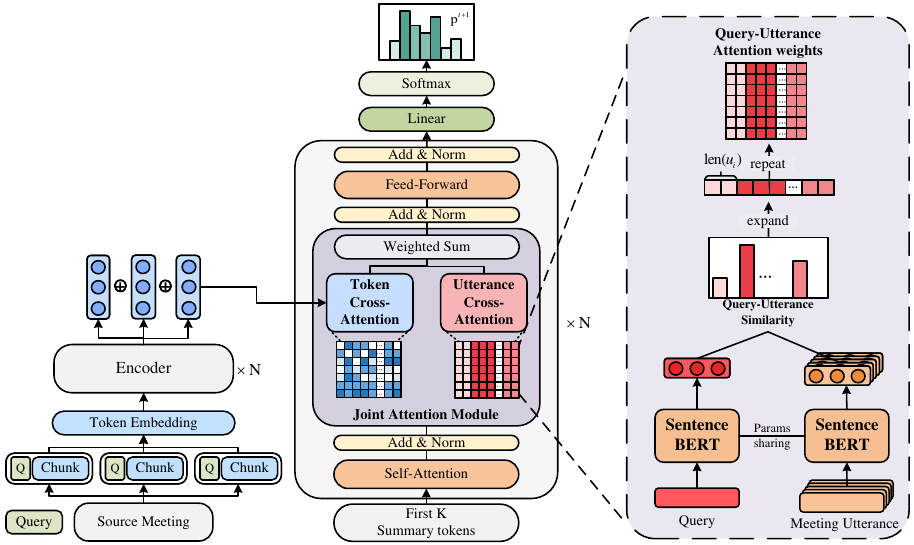}%
	\caption{The overall model structure}
	\label{fig2}
	\vspace{-0.1cm}
\end{figure*}

Compared with the generic summary, QFMS is faced with the challenge of generating a summary related to the query. 
Previous works typically concatenate the query with meeting transcripts and implicitly model the query relevance only at the token level with attention mechanism.
However, it is not enough to highlight the key parts related to the query in long meeting transcripts.
As shown in Figure \ref{fig1}, the user is wondering about the disagreement between the industrial designer and the marketing regarding the product's future trend. 
The summary generated by previous work incorrectly focuses on discussions about product shape.
This is likely due to the fact that the information related to the query will be diluted with the increase in meeting length. The previous original transformer-based model does not explicitly model the correlation with the query, but only models the correlation between tokens with the attention mechanism, which does not contain the complete semantic information as utterance.
Existing work \cite{zhong-etal-2021-qmsum, vig-etal-2021-exploring} takes a select-then-generate approach to process long input and feed the most relevant parts to the generation model.
A major drawback faced by the two-stage approach is error propagation, some parts which are incorrectly filtered out don't appear in the generation stage at all, which even aggravates the error that the generated summaries are not related to the query. 
Hence, we believe it makes sense to explore an end-to-end approach that explicitly integrates Query-Utterance relevance in the generation process.

To address the above challenges, in this paper, we propose a query-aware framework by jointly modeling query relevance with token and utterance. 
First, we use an additional dense retrieval module to calculate the similarity between the query and each meeting utterance.
Second, we adopt an end-to-end model and replace the origin cross attention with token cross attention and Query-Utterance cross attention. 
With a joint training framework with the query relevance of different granularity, the model can generate summaries that are more related to the query.

To summarize, our contributions are as follows:
(1) Our work demonstrates the effectiveness of the Query-Utterance relevance in generating a summary more related to the query.
(2) We propose a query-aware framework by jointly modeling query relevance with both token and utterance. 
(3) Experimental results show that our proposed model achieves new state-of-the-art performance.


\section{METHOD}
The architecture of our method is illustrated in Figure \ref{fig2}. 
Our model mainly consists of three components: 
Input chunking is designed for long meeting transcripts. 
Query-Utterance relevance module is used to calculate the similarity between each utterance and the query. 
Then by end-to-end training of joint attention module, the summary is generated.
\subsection{Input Chunking}
In order to feed the whole meeting into the model, we follow the preprocessing step \cite{vig-etal-2021-exploring} to split the source meeting into several fixed-length overlapping chunks, which are 50\% overlapping.
As shown in Figure \ref{fig2},
each chunk is concatenated with the query in front of it. 
These chunks are fed into a standard Transformer encoder. 
Then we concatenate these encodings to get a single embedding sequence and pass it to the modified decoder. 

\subsection{Query-Utterance Relevance}
A typical meeting consists of multiple interactions with multiple members.
We consider one statement by a participant in the meeting transcript as an utterance, the meeting transcript $U$ is the form of a sequence of utterances $(u_1, u_2, ..., u_n)$.
To get the Query-Utterance relevance, we calculate the similarity between the embeddings between them.

SBERT \cite{reimers-gurevych-2019-sentence} is proposed for dense retrieval, which adopts the siamese network architecture to encode the sentence pair separately and uses cosine-similarity to get the similarity between them. 

For given query $q$ and utterance $u$, we use SBERT to encode them to sentence embedding $e_q$ and $e_u$. 
The query $q$ contains $l_q$ tokens $q=(w_i, w_2, ..., w_{l_q})$, the utterance $u$ contains $l_u$ tokens $u_i=(t_1, t_2, ..., t_{l_u})$.

SBERT contains a pre-trained BERT for sentence pair regression task and a mean-pooling layer.
The sentence embedding $e_q$ and $e_u$ are as follows:

\begin{equation}
    e_q=\frac{1}{l_q}\sum_{i=1}^{l_q}(e_{q_i});\ \ e_u=\frac{1}{l_u}\sum_{i=1}^{l_u}(e_{u_i})
\end{equation}
\begin{equation}
    (e_{q_1}, e_{q_2},..., e_{q_l})=BERT(w_i, w_2, ..., w_{l_q})
\end{equation}
\begin{equation}
    (e_{u_1}, e_{u_2},..., e_{u_l})=BERT(t_i, t_2, ..., t_{l_u})
\end{equation}
The similarity of query $q$ and utterance $u$ is measured with cosine similarity.
\begin{equation}
    s=cosine\raisebox{0mm}{-}sim(e_q, e_u)
\end{equation}

\subsection{Joint Attention Module}
After getting the Query-Utterance relevance, 
we propose a joint attention module to fuse query relevance at different granularities and incorporate the relevance with attention mechanism in the generation process.
Since the concatenation of the query and meeting transcript is used as the encoder's input, the query and utterance interact implicitly through the attention mechanism at the token level.

While calculating similarity, the query and utterance interact explicitly through the Query-Utterance relevance module at the sentence level. 
We believe that paying attention to the different granularities of correlation does help to better generate summaries related to the query. 

In this module, query relevance at different granularities are fused by joint attention mechanism. 
Then we incorporate the Query-Utterance relevance in the generation process to generate summaries more related to the query.

Given a query $q$ and utterances $U=(u_1, u_2, ..., u_n)$, for each utterance $u_i$, we can get the similarity $s_i$ of them according to the Query-Utterance relevance module. 
The Query-Utterance relevance is $S=(s_1, s_2, ... ,s_n)$. As shown in Figure \ref{fig2}, to ensure the tokens in the same utterance should be given the same Query-Utterance relevance score, if $u_i$ contains $l_i$ tokens, $s_i$ should repeat $l_i$ times. Then all the Query-Utterance relevance score will be expand to the same shape with Encoder-Decoder cross attention weights $\alpha$, which is called Query-Utterance attention weights $A_{Q\raisebox{0mm}{-}U}$.
Finally, we apply weighted-sum to Encoder-Decoder cross attention weights and Query-Utterance attention weights to get the $l$-th transformer decoder layer's attention weights $\alpha^l$.
\begin{equation}
    Attention(Q,K,V,A_{Q\raisebox{0mm}{-}U})=softmax(\alpha^l\cdot V)
\end{equation}
\begin{equation}
    \label{eq6}
    \alpha^l = (1-\beta)\cdot \frac{QK^T}{\sqrt{d}}+\beta \cdot A_{Q\raisebox{0mm}{-}U}
\end{equation}
where $d$ is the dimension of the query matrix $Q$ and key matrix $K$, $\beta$ is a weight hyperparameter.

\section{EXPERIMENTS}

\subsection{Setup}
\subsubsection{Implementation Details}
\label{sec:appendixa}
Models are implemented using the PyTorch framework. 
We use the pre-trained BART\footnote{The checkpoint is “facebook/bart-large”.} from the Transformers 
library as the base abstractive model, and dense retrieval model RoBERTa\footnote{The checkpoint is "sentence-transformers/all-roberta-large-v1".} from sentence-transformers  \cite{reimers-gurevych-2019-sentence} library as the sentence similarity model.

All experiments are conducted on 2 NVIDIA RTX 3090 GPUs(24G memory). 
Each model is trained for 10 epochs, the average running time is around 12 hours.
Weight hyperparameter $\beta$ is 0.6 in equation \ref{eq6}. Chunk size is 512, max number of chunks is 32. 
Learning rate is 5e-6.

Models were evaluated using the ROUGE metrics \cite{lin-2004-rouge} in the SummEval toolkit \cite{fabbri-etal-2021-summeval} and each pair of results was subjected to t-test to confirm the effectiveness of our method.


\subsubsection{Datasets Details}
\label{sec:appendixb}
\textbf{QMSum} \cite{zhong-etal-2021-qmsum} is a query-focused meeting summarization dataset, which consists of 1,808 query-summary pairs over 232 meetings from product design, academic, and political committee meetings. 
QMSum also contains additional manual annotations such as topic segmentation and relevant spans related to the reference summary.


\subsubsection{Baselines Details}
\label{sec:appendixc}
We compare the proposed method with several baselines. \textbf{TextRank} \cite{mihalcea-tarau-2004-textrank} is a graph-based ranking model to extract key information. 
\textbf{PGN} \cite{see-etal-2017-get} uses pointer mechanism to copy tokens from source texts.
\textbf{BART} \cite{lewis-etal-2020-bart} is an encoder-decoder Transformer model pre-trained using a denoising objective.
\textbf{HMNet} \cite{zhu-etal-2020-hierarchical} utilizes a hierarchical attention structure and cross-domain pre-training for meeting summarization.
\textbf{SUMM$^N$} \cite{zhang2021summn} is a multi-stage summarization framework for long input dialogues and documents.
\textbf{Point Network+PGNet} and \textbf{Point Network+BART} \cite{zhong-etal-2021-qmsum} adopt a two-stage approach of locate-then-summarize for long meeting summarization. 
\textbf{SBERT+BART} is another two-stage baseline proposed in this paper. 
\textbf{Longformer} \cite{Beltagy2020Longformer} replaces the quadratic self-attention mechanism with a combination of local attention and sparse global attention.
\textbf{SEGENC} \cite{vig-etal-2021-exploring} splits the meeting into fixed-length segments, then concatenates the segment embeddings into a single sequence and passes it to the decoder. Without Query-Utterance attention, our model can be regarded as the replication of SEGENC.


\begin{table}
\centering
\resizebox{1.02\columnwidth}{!}{
\begin{tabular}{lccc}
\toprule
\textbf{Models} & \textbf{ROUGE-1} & \textbf{ROUGE-2} & \textbf{ROUGE-L}\\
\midrule
TextRank \cite{mihalcea-tarau-2004-textrank}        &   16.27   &   2.69    &   15.41   \\
PGNet \cite{see-etal-2017-get}           &   28.74   &   5.98    &   25.13   \\
BART \cite{lewis-etal-2020-bart}            &   29.20   &   6.37    &   25.49   \\
HMNet \cite{zhu-etal-2020-hierarchical}  &   32.29   &   8.67    &   28.17   \\
SUMM$^N$ \cite{zhang2021summn}                   &    34.03  &   9.28    &   29.48   \\
DYLE \cite{mao-etal-2022-dyle}   &    34.42   &   9.71    &   30.10   \\
DialogLM \cite{Zhong2022DialogLMPM}    &   33.69   &   9.32    &   30.01   \\
\midrule
\hspace{0.5cm}
two-stage model  \\
Point Network + PGNet \cite{zhong-etal-2021-qmsum}  &   31.37   &   8.47    &   27.08   \\
Point Network + BART \cite{zhong-etal-2021-qmsum}   &   31.74   &   8.53    &   28.21   \\
SBERT + BART    &   34.21   &   10.37   &   29.80   \\
\midrule
\hspace{0.5cm}
end-to-end model  \\
Longformer \cite{Beltagy2020Longformer}  &   34.18   &   10.32   &   29.95   \\
SEGENC \cite{vig-etal-2021-exploring}          &   37.05   &   13.03   &   32.62   \\
\midrule
\textbf{Ours}   &   \textbf{37.99}   &   \textbf{13.66}   &   \textbf{33.36}   \\
Ours(w/o Q-U Attention)  &   37.03   &   12.93   &   32.47   \\
\bottomrule
\end{tabular}
}
\caption{
\label{tab1}
ROUGE-F1 scores for different models on QMSum dataset. 
}
\end{table}
\vspace{-0.3cm}
\subsection{Results \& Analysis}
The performances of our method and baselines are summarized in Table \ref{tab1}.
Experimental results show that our method significantly outperforms the baselines (p \textless 0.05) and achieves new state-of-the-art results on QMSum dataset.

To feed the input into the fixed-length model like BART, we have to truncate the source meeting to fixed length.
Experimental results show that truncation operations lead to decreased performance.
For generic summarization, lead bias reduces the impact of truncation operations on summarization. 
But for QFMS, highlights related to the query may appear in different parts of the meeting transcripts. Therefore QFMS is particularly affected by truncation operations.

Besides, We observe that in general end-to-end models perform better than the two-stage models once the long input problem is solved. 
It shows that benefiting from attention mechanism, existing end-to-end models have a certain ability to concentrate on the parts related to the query and generate summaries. 
Based on previous end-to-end models, our model achieves a higher ROUGE score by incorporating Query-Utterance relevance in the generation process.

To verify the effectiveness of Query-Utterance relevance, we conduct an ablation experiment.
Our model significantly outperformed the model without Query-Utterance relevance (p \textless 0.05).
Experimental results show that the model without Query-Utterance relevance reduces 0.96 ROUGE-1, 0.73 ROUGE-2, 0.89 ROUGE-L scores, which demonstrates the effectiveness of incorporating the  Query-Utterance relevance in end-to-end models.

\begin{table}
\centering
\resizebox{0.85\columnwidth}{!}{
\begin{tabular}{lccc}
\toprule
\textbf{Models} & \textbf{ROUGE-1} & \textbf{ROUGE-2} & \textbf{ROUGE-L} \\
\midrule
BART & 10.10 & 1.84 & 9.27 \\
SBERT+BART & 13.21 & 3.36 & 12.25 \\
SEGENC & 14.32 & 5.18 & 13.74 \\
Ours & \textbf{16.06} & \textbf{5.90} & \textbf{15.33} \\
\bottomrule
\end{tabular}
}
\caption{
\label{tab2}
ROUGE recall score between annotated query-relevant meeting utterances and the summary.
}
\end{table}

\subsection{Query-Relevant Information Recall}
QMSum dataset contains annotations of utterances related to the query.
We use ROUGE recall score between annotated query-relevant meeting utterances and the summary to evaluate the ability to capture the information related to the query.
Experimental result is shown in Table \ref{tab2}. 
It shows our model achieves significant recall score improvement compared with the model without Query-Utterance relevance, 
which demonstrates the effectiveness of the Query-Utterance relevance in highlight the key information related to query (p \textless 0.01).


\begin{table}
\centering
\resizebox{0.54\columnwidth}{!}{
\begin{tabular}{lccc}
\toprule
\textbf{Models} & \textbf{Flu.} & \textbf{QR.} & \textbf{FC.} \\
\midrule
Gold & 4.88 & 4.90 & 4.92 \\
BART & 4.48 & 3.78 & 3.64 \\
SEGENC & 4.24 & 3.96 & 4.02 \\
Ours & 4.32 & \textbf{4.44} & \textbf{4.28} \\
\bottomrule
\end{tabular}
}
\caption{
\label{tab3}
Human evaluation on Fluency (Flu.), Query Relevance (QR.) and Factual Consistency (FC.) for QMSum.
}
\end{table}

\subsection{Human Evaluation}
We further conduct a manual evaluation to assess
the models. We randomly select 50 samples from QMSum and ask 10 professional linguistic evaluators to score the ground truth and summaries generated by 3 models according to 3 metrics: fluency,
query relevance and factual consistency. Each metric is rated by 3 evaluators from 1 (worst) to 5 (best) and the scores for each summary are averaged.

As shown in Table \ref{tab3}. With Query-Utterance attention, our model significantly outperforms original BART and SEGENC on query relevance, which shows the effectiveness of joint training of query-token and token-utterance attention. 
Results on factual consistency scores also demonstrate the contribution of modeling query relevance to improving factual consistency. 
And it also shows the fused attention mechanism results in a slight decrease in fluency scores.


\section{Conclusion}
This paper presents that query relevance at different granularities complements each other and contributes to QFMS.
To take different granularities into consideration, we propose a query-aware framework by jointly modeling token and utterance,
which models explicit Query-Utterance relevance and employs a joint attention module to incorporate the relevance in the generation process with attention mechanism. 
Experiments demonstrate the effectiveness of Query-Utterance relevance in generating summaries more related to the query.
\small
\bibliographystyle{IEEEbib}
\bibliography{refs}

\begin{thebibliography}{10}

\bibitem{litvak-vanetik-2017-query}
Marina Litvak and Natalia Vanetik,
\newblock ``Query-based summarization using {MDL} principle,''
\newblock in {\em Proceedings of the {M}ulti{L}ing 2017 Workshop on
  Summarization and Summary Evaluation Across Source Types and Genres},
  Valencia, Spain, Apr. 2017, pp. 22--31, Association for Computational
  Linguistics.

\bibitem{baumel2018query}
Tal Baumel, Matan Eyal, and Michael Elhadad,
\newblock ``Query focused abstractive summarization: Incorporating query
  relevance, multi-document coverage, and summary length constraints into
  seq2seq models,''
\newblock {\em arXiv preprint arXiv:1801.07704}, 2018.

\bibitem{laskar-etal-2020-wsl}
Md~Tahmid~Rahman Laskar, Enamul Hoque, and Jimmy~Xiangji Huang,
\newblock ``{WSL}-{DS}: Weakly supervised learning with distant supervision for
  query focused multi-document abstractive summarization,''
\newblock in {\em Proceedings of the 28th International Conference on
  Computational Linguistics}, Barcelona, Spain (Online), Dec. 2020, pp.
  5647--5654, International Committee on Computational Linguistics.

\bibitem{xu-lapata-2020-coarse}
Yumo Xu and Mirella Lapata,
\newblock ``Coarse-to-fine query focused multi-document summarization,''
\newblock in {\em Proceedings of the 2020 Conference on Empirical Methods in
  Natural Language Processing (EMNLP)}, Online, Nov. 2020, pp. 3632--3645,
  Association for Computational Linguistics.

\bibitem{xu-lapata-2021-generating}
Yumo Xu and Mirella Lapata,
\newblock ``Generating query focused summaries from query-free resources,''
\newblock in {\em Proceedings of the 59th Annual Meeting of the Association for
  Computational Linguistics and the 11th International Joint Conference on
  Natural Language Processing (Volume 1: Long Papers)}, Online, Aug. 2021, pp.
  6096--6109, Association for Computational Linguistics.

\bibitem{zhong-etal-2021-qmsum}
Ming Zhong, Da~Yin, Tao Yu, Ahmad Zaidi, Mutethia Mutuma, Rahul Jha,
  Ahmed~Hassan Awadallah, Asli Celikyilmaz, Yang Liu, Xipeng Qiu, and Dragomir
  Radev,
\newblock ``{QMS}um: A new benchmark for query-based multi-domain meeting
  summarization,''
\newblock in {\em Proceedings of the 2021 Conference of the North American
  Chapter of the Association for Computational Linguistics: Human Language
  Technologies}, Online, June 2021, pp. 5905--5921, Association for
  Computational Linguistics.

\bibitem{vig-etal-2021-exploring}
Jesse Vig, Alexander~R. Fabbri, Wojciech Kry{\'s}ci{\'n}ski, Chien-Sheng Wu,
  and Wenhao Liu,
\newblock ``Exploring neural models for query-focused summarization,'' 2021.

\bibitem{reimers-gurevych-2019-sentence}
Nils Reimers and Iryna Gurevych,
\newblock ``Sentence-{BERT}: Sentence embeddings using {S}iamese
  {BERT}-networks,''
\newblock in {\em Proceedings of the 2019 Conference on Empirical Methods in
  Natural Language Processing and the 9th International Joint Conference on
  Natural Language Processing (EMNLP-IJCNLP)}, Hong Kong, China, Nov. 2019, pp.
  3982--3992, Association for Computational Linguistics.

\bibitem{lin-2004-rouge}
Chin-Yew Lin,
\newblock ``{ROUGE}: A package for automatic evaluation of summaries,''
\newblock in {\em Text Summarization Branches Out}, Barcelona, Spain, July
  2004, pp. 74--81, Association for Computational Linguistics.

\bibitem{fabbri-etal-2021-summeval}
Alexander~R. Fabbri, Wojciech Kry{\'s}ci{\'n}ski, Bryan McCann, Caiming Xiong,
  Richard Socher, and Dragomir Radev,
\newblock ``{S}umm{E}val: Re-evaluating summarization evaluation,''
\newblock {\em Transactions of the Association for Computational Linguistics},
  vol. 9, pp. 391--409, 2021.

\bibitem{mihalcea-tarau-2004-textrank}
Rada Mihalcea and Paul Tarau,
\newblock ``{T}ext{R}ank: Bringing order into text,''
\newblock in {\em Proceedings of the 2004 Conference on Empirical Methods in
  Natural Language Processing}, Barcelona, Spain, July 2004, pp. 404--411,
  Association for Computational Linguistics.

\bibitem{see-etal-2017-get}
Abigail See, Peter~J. Liu, and Christopher~D. Manning,
\newblock ``Get to the point: Summarization with pointer-generator networks,''
\newblock in {\em Proceedings of the 55th Annual Meeting of the Association for
  Computational Linguistics (Volume 1: Long Papers)}, Vancouver, Canada, July
  2017, pp. 1073--1083, Association for Computational Linguistics.

\bibitem{lewis-etal-2020-bart}
Mike Lewis, Yinhan Liu, Naman Goyal, Marjan Ghazvininejad, Abdelrahman Mohamed,
  Omer Levy, Veselin Stoyanov, and Luke Zettlemoyer,
\newblock ``{BART}: Denoising sequence-to-sequence pre-training for natural
  language generation, translation, and comprehension,''
\newblock in {\em Proceedings of the 58th Annual Meeting of the Association for
  Computational Linguistics}, Online, July 2020, pp. 7871--7880, Association
  for Computational Linguistics.

\bibitem{zhu-etal-2020-hierarchical}
Chenguang Zhu, Ruochen Xu, Michael Zeng, and Xuedong Huang,
\newblock ``A hierarchical network for abstractive meeting summarization with
  cross-domain pretraining,''
\newblock in {\em Findings of the Association for Computational Linguistics:
  EMNLP 2020}, Online, Nov. 2020, pp. 194--203, Association for Computational
  Linguistics.

\bibitem{zhang2021summn}
Yusen Zhang, Ansong Ni, Ziming Mao, Chen~Henry Wu, Chenguang Zhu, Budhaditya
  Deb, Ahmed~H Awadallah, Dragomir Radev, and Rui Zhang,
\newblock ``Summ\^{} n: A multi-stage summarization framework for long input
  dialogues and documents,''
\newblock in {\em ACL 2022}, 2022.

\bibitem{Beltagy2020Longformer}
Iz~Beltagy, Matthew~E. Peters, and Arman Cohan,
\newblock ``Longformer: The long-document transformer,''
\newblock {\em arXiv:2004.05150}, 2020.

\bibitem{mao-etal-2022-dyle}
Ziming Mao, Chen~Henry Wu, Ansong Ni, Yusen Zhang, Rui Zhang, Tao Yu,
  Budhaditya Deb, Chenguang Zhu, Ahmed Awadallah, and Dragomir Radev,
\newblock ``{DYLE}: Dynamic latent extraction for abstractive long-input
  summarization,''
\newblock in {\em Proceedings of the 60th Annual Meeting of the Association for
  Computational Linguistics (Volume 1: Long Papers)}, Dublin, Ireland, May
  2022, pp. 1687--1698, Association for Computational Linguistics.

\bibitem{Zhong2022DialogLMPM}
Ming Zhong, Yang Liu, Yichong Xu, Chenguang Zhu, and Michael Zeng,
\newblock ``Dialoglm: Pre-trained model for long dialogue understanding and
  summarization,''
\newblock in {\em AAAI}, 2022.

\end{thebibliography}

\end{document}